\documentclass[conference]{IEEEtran}
\IEEEoverridecommandlockouts
\usepackage{cite}
\usepackage{amsmath,amssymb,amsfonts}
\usepackage{algorithmic}
\usepackage{graphicx}
\usepackage{textcomp}
\usepackage{xcolor}
\def\BibTeX{{\rm B\kern-.05em{\sc i\kern-.025em b}\kern-.08em
    T\kern-.1667em\lower.7ex\hbox{E}\kern-.125emX}}
\begin{document}

\title{Spatio-Temporal Human Action Recognition Model with Flexible-interval Sampling and Normalization\\

}

\author{\IEEEauthorblockN{ Yuke Yang\textsuperscript}
\IEEEauthorblockA{\textit{} \\
yangyuke@zju.edu.cn}
}

\maketitle

\begin{abstract}
Human action recognition\cite{Zhou_2018_CVPR} is a well-known computer vision and pattern recognition task of identifying which action a man is actually doing. Extracting the keypoint information of a single human with both spatial and temporal features of action sequences plays an essential role to accomplish the task\cite{2018_CVPR, cao2018openpose, simon2017hand, cao2017realtime, wei2016cpm,fang2017rmpe,xiu2018poseflow}. In this paper, we propose a human action system for Red-Green-Blue(RGB) input video with our own designed module. Based on the efficient Gated Recurrent Unit(GRU)\cite{pmlr-v63-gao30, chung2014empirical, 8053243} for spatio-temporal feature extraction, we add another sampling module and normalization module to improve the performance of the model in order to recognize the human actions. Furthermore, we build a novel dataset with a similar background and discriminative actions for both human keypoint prediction and behavior recognition. To get a better result, we retrain the pose model with our new dataset to get better performance. Experimental results demonstrate the effectiveness of the proposed model on our own human behavior recognition dataset and some public datasets.
\end{abstract}


\section{Introduction}
Human action recognition has been a fundamental and challenging task in computer vision over these years. Many related areas such as intelligent video surveillance\cite{Wang:2013:IMV:2397196.2397216}, human-computer interaction\cite{Karray_human-computerinteraction:}, video summary and understanding\cite{Miki_Haseyama2013}, and smart device action recognition\cite{Zhou_2018_CVPR}  are prevalent due to the development of human action recognition. Most of the current methods focus on three types of input -- RGB, optical and skeleton videos\cite{Song:2017:ESA:3298023.3298186,8237377,Liu_2017_CVPR,8026287}. Among these three types, RGB is most closed to the practical usage of real-time human recognition techniques because it is the direct source of the data. For RGB videos, it has been proved that extracting discriminative spatial-temporal features\cite{10.1007/978-3-642-53932-9_52} would effectively model the spatial and temporal evolutions of different actions.\\

Even though the current human action recognition task has been popular, public dataset still has two problems. The first one is about the background. For example, UCF-101\cite{abs-1212-0402} has lots of action types that differ from each other a lot, but the backgrounds of these images also vary a lot. So for any algorithm with high precision on UCF-101, it may classify video based on the background instead of recognizing the behaviors precisely.\cite{10.1145/2939672.2939778} Meanwhile, for some datasets containing similar backgrounds, their action classification is too closed even for human-beings. For example, for the HON4D Action dataset\cite{Oreifej_2013_CVPR}, lifting the cup and drinking water are too resembled to distinguish even for humans so we can expect it is difficult for the model to hold a high accuracy there. The second one is not as serious as the first one, but it still needs to be noticed here -- the localization problem. Some models own good performance on certain datasets such as the UCLA-Northwestern Action Dataset\cite{Wang_2014_CVPR} because most of their actions are taken place in the same areas of videos. The models can be sensitive to the actions in these areas but not actions on the other areas. As a result, a dataset containing similar backgrounds and discriminative actions is in demand.\\

Three kinds of methods are prevalent through the whole human recognition on RGB task: two-stream method\cite{NIPS2014_5353,Feichtenhofer_2016_CVPR}, three-dimension convolution method\cite{AAAI1817205,801497123211,Tran_2015_ICCV,2017arXiv171108200D} and spatio-temporal types of method\cite{10.1007/978-3-319-46487-9_50,Zhang_2017_ICCV}. However, the first and second methods suffer from severe shortcomings. Two-stream methods extensively developed these years, and some of them achieved a relatively high precision on public datasets such as UCF-101. But according to some research\cite{10.1145/2939672.2939778}, most of the two-stream methods depend largely on the background of the RGB videos so this type of method did not recognize the action but background instead. Some modern work\cite{10.1007/978-3-030-12939-2_20} have already shown that the precision of the two-stream method has fallen down on the datasets with a similar background, let alone the two-stream method costs a lot of computing resources. Three-dimension convolution methods such as T3D\cite{2017arXiv171108200D} and C3D\cite{AAAI1817205} can extract some spatio-temporal feature but they are limited. They do not concentrate on adequate length of the video so they do not have such high precision on some datasets such as HMDB51\cite{Kuehne11}. Consequently, for the RGB videos, the spatio-temporal method would be most useful.\\

However, not all aspects are perfect for the current spatio-temporal methods\cite{10.1007/978-3-319-46487-9_50,Zhang_2017_ICCV}. Previous work has made some achievements in human action recognition but their methods own two drawbacks. The first one is the results would be misleading if the original video contains sufficient time interference factor\cite{10.1007/978-3-030-12939-2_20,abs-1212-0402}. For example, a 10-second falling video may contain 4-second standing time, 3-second falling time and 3-second lying time so that the algorithm would classify the video as standing or lying, which would be unexpected because it is more important to focus on the rapid change in the video. The second one is for some of these spatio-temporal methods with sequential sampling\cite{8297027}, each sample may be misleading because it may not cover the whole action. Some actions such as archery in UCF-101 only cover 4 frames, but the whole video is much longer. 20 sequential frames of sampling in these methods must be insufficient. According to these two difficulties, a sampling module can be effective.\\

Faced with these challenges, we build up our own dataset and propose a human action recognition system. Specifically for our own dataset, we count certain common behaviors with almost the same background in similar places and light conditions with only one person inside each video. For our model, it recognizes the human behaviors after extracting the bounding box from yolo-version three\cite{Redmon2018YOLOv3AI} and keypoint information extracting from pose model\cite{2018_CVPR, cao2018openpose, simon2017hand, cao2017realtime, wei2016cpm,fang2017rmpe,xiu2018poseflow}. This paper integrates the pose model and the latter spatio-temporal network. To solve the challenges above, this paper proposes a solution to deal with the keypoint information with flexible-interval sampling and normalization module inside the model in both training and verifying process. The flexible-interval sampling and normalization both work for increasing our model generalization -- sampling works on the time and normalization works on the space. Our work replaces the conventional frequently-used LSTM\cite{Hochreiter,Sak2014LongSM} with GRU\cite{pmlr-v63-gao30, chung2014empirical, 8053243}, because the latter one demonstrates better performance on this type of work and can have fewer parameters. This method not only assists the model to be more precise but generate more training set. \\

To summarize, we have made four main contributions:\\
\begin{itemize}
    \item we build our own dataset with a similar background and discriminative actions.
    \item we add a unique sampling module to improve the accuracy of the model.	\item we add a unique normalization method to increase generalization on various kinds of videos.
    \item we retrain the pose model with our own labeled data to improve the accuracy of the pose model.
\end{itemize}

The remainder of this paper is structured as follows. Section II presents how we collect dataset and Section III shows the composition of our model. Section IV describes how we conduct experiments. Results and discussions are shown in Section V. Conclusion and future research directions are given in Section VI.\\

\section{Dataset}
In this section, we first introduce the details about collecting our action dataset named Spatio-Temporal Human Dataset(STH-Dataset) consisting of multiple actions including standing, sitting, walking, waving, kicking, falling down and other unrecognized behaviors. The eventual category may contain the videos with an incomplete body which leads to unidentifiable action. Figure 1 shows examples of our dataset.\\

\begin{figure}[htbp]
\centerline{\includegraphics[width = 8cm]{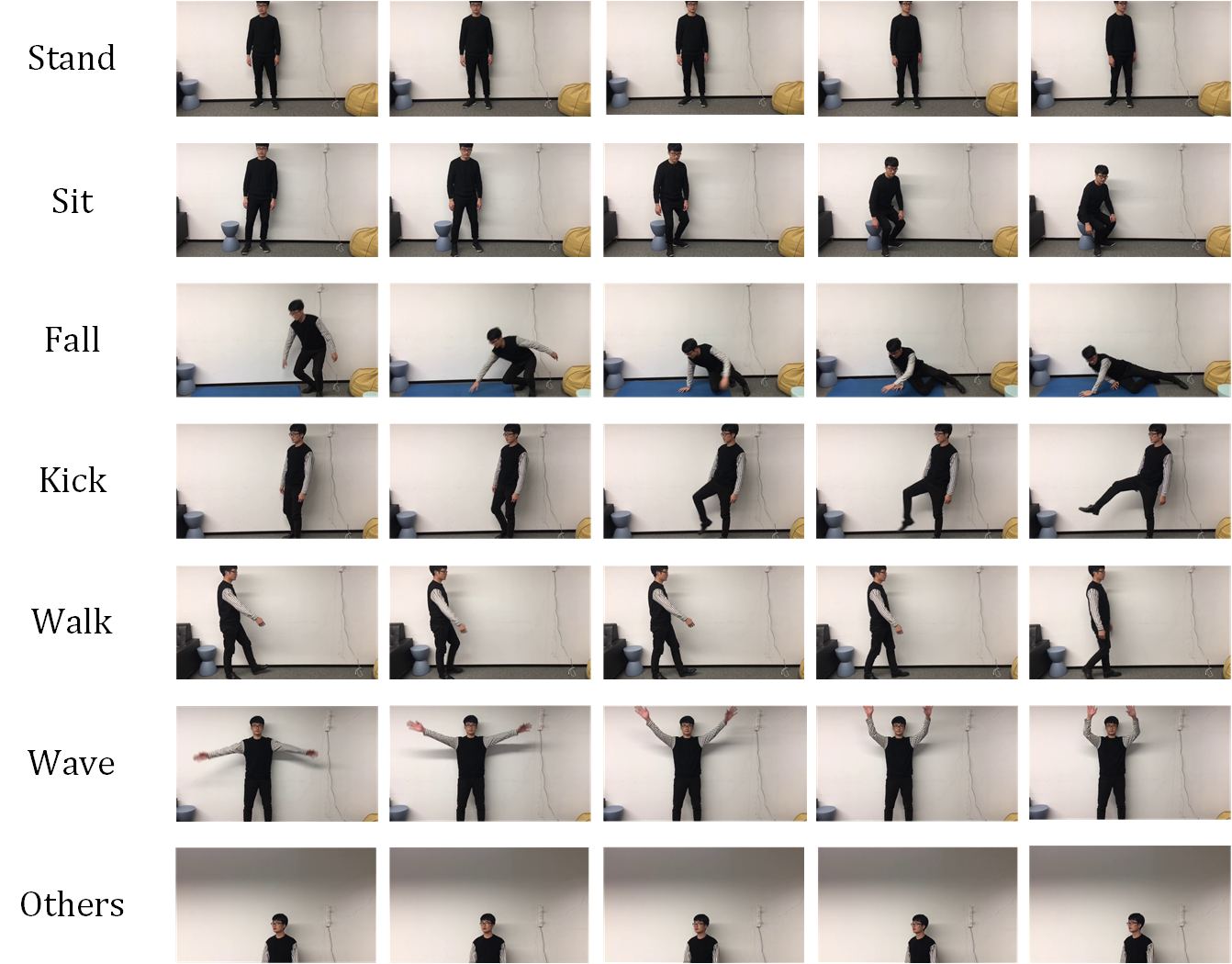}}
\caption{Example of Dataset. STH-Dataset sampled from our collected human action behavior dataset. The original source is videos and we display five sequential frames to represent the whole behavior. }
\label{fig}
\end{figure}

\subsection{Collecting Method}
We collect our data\cite{doi:10.1177/1729881417709079} in three kinds of scenes: the light office with adequate decorations, the light office with few decorations and the dark office with even no decoration. Generally, these backgrounds are semblable but they all have nuances between themselves so that the perfect performance on our dataset can prove that the model really focuses on actions. Specifically, each scene consists of 9 cameras: top-front left corner, top-front right corner, top-back left corner, top-back right corner, bottom-front left corner, bottom-front right corner, bottom-back left corner, bottom-back right corner and one more camera with the same level as the normal human-beings' height(1.35 meters when sitting on the chair). These cameras take videos of the same action from different angles. \\

The cameras are named DS-2CD3T45-I3 camera produced by Haikang Inc. . One mobile phone is connected to only one camera to record each video and store them on the SD card. Finally, we clear up the dataset and store them according to the category. \\

\subsection{Composition}
According to table 1, our dataset consists of 7 kinds of actions including standing, sitting, moving, waving, falling down and other unrecognized behaviors with different persons. The total number of our dataset has already reached 9604. The detailed information can be seen in table 1. All of the videos are marked with 18 keypoint information and the type of action.\\
\begin{table}[htbp]
    \centering
    \caption{The Number of Each Action in STH-Dataset}
    \begin{tabular}{|p{3cm}||p{3cm}|}
    \hline
    \textbf{Action Type}&\textbf{Number}\\
    \hline
    Wave& 879\\
    Walk& 321\\
    Stand&1808\\
    Fall&3851\\
    Kick&1234\\
    Sit&1136\\
    Others&375\\
    \hline
    \textbf{Total}&9604\\
    \hline
    \multicolumn{2}{l}{$^{\mathrm{a}}$ The table contains a specific number of each category of action.}
    \end{tabular}
    \label{tab:my_label}
\end{table}

\section{Our Proposed Model}
In this section, we describe our proposed Spatio-Temporal Human Action Model in detail.
We first introduce the background information of the Pose model\cite{2018_CVPR, cao2018openpose, simon2017hand, cao2017realtime, wei2016cpm,fang2017rmpe,xiu2018poseflow} and the Gated Recurrent Unit technique\cite{pmlr-v63-gao30, chung2014empirical, 8053243}. Both of them are in the leading position of the current method. Second, we notice how our model is designed based on these two kinds of techniques with our own sampling module.\\
\subsection{Background information}
\paragraph{Pose}
Modern techniques for multiple human keypoint estimation have been various and effective. We utilize these mature models to be trained on the combination of our dataset and the public dataset -- coco dataset \cite{10.1007/978-3-319-10602-1_48}. Specifically, we use existed model called AlphaPose\cite{fang2017rmpe,xiu2018poseflow} to train a pose model with the input of RGB image and output the keypoint information of human beings. Openpose\cite{2018_CVPR, cao2018openpose, simon2017hand, cao2017realtime, wei2016cpm}  is another similar model but here we choose the AlphaPose because of its excellent performance. \\ 

Here is a brief explanation of how the AlphaPose model works. The AlphaPose model is a Regional Multi-person Pose Estimation(RMPE)\cite{Fang_2017_ICCV} work, which means it can predict keypoints in one image of multiple persons utilizing the proposal method\cite{NIPS2015_5638}. Generating the human bounding boxes by the human detector, these images with different bounding boxes are fed into the Symmetric Spatial Transformer Network(SSTN)\cite{NIPS2015_5854} and Single-Person Pose Estimator(SPPE) module, and they would be transferred into the pose proposals automatically. SSTN is useful to extract a high-quality single person region from an inaccurate bounding box and SPPE is useful to extract keypoint information from a single-human image. The pose proposal selection module would finish the final selection. With Non-Maximum-Suppression(NMS)\cite{Cai_2019_CVPR} method, the final output keypoints would be filtered out.\\ 

This work won't introduce the details of this model. In this section, we utilize the AlphaPose model to predict human keypoints.  Besides the original coco\cite{10.1007/978-3-319-10602-1_48} dataset to train the pose model, we add one another public dataset downloading from the AI challenge contest website(https://challenger.ai/) and our own dataset to supplement the human keypoint dataset. The keypoints criterion is identical to the coco human keypoints -- 18 keypoints for each human. The details would be in the experiment section.\\
\paragraph{Gated Recurrent Unit}
Better than long short-term memory (LSTM)\cite{Hochreiter,Sak2014LongSM}, gated recurrent units(GRU)\cite{pmlr-v63-gao30, chung2014empirical, 8053243} technique is extensively used in recurrent neural networks these years. It lacks an output gate compared to the LSTM, consequently, most of the GRU has fewer parameters than LSTM. It has been proved that for some large datasets, GRU works as well as the LSTM. However, for some small datasets, GRU even performs more effectively.\\

For this model, the GRU model utilization details are as follows\cite{69e088c8129341ac89810907fe6b1bfe} modulating the flow of information inside the unit without having separate memory cells.\\
\begin{figure}[htbp]
\centerline{\includegraphics{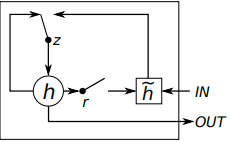}}
\caption{Gated Recurrent Unit Network Illustration. The r gate sets how much the gate would forget about its input information and the z gate would control which input it would depend on.}
\label{fig}
\end{figure}

\begin{figure*}[htbp]
\centerline{\includegraphics[width = 18cm]{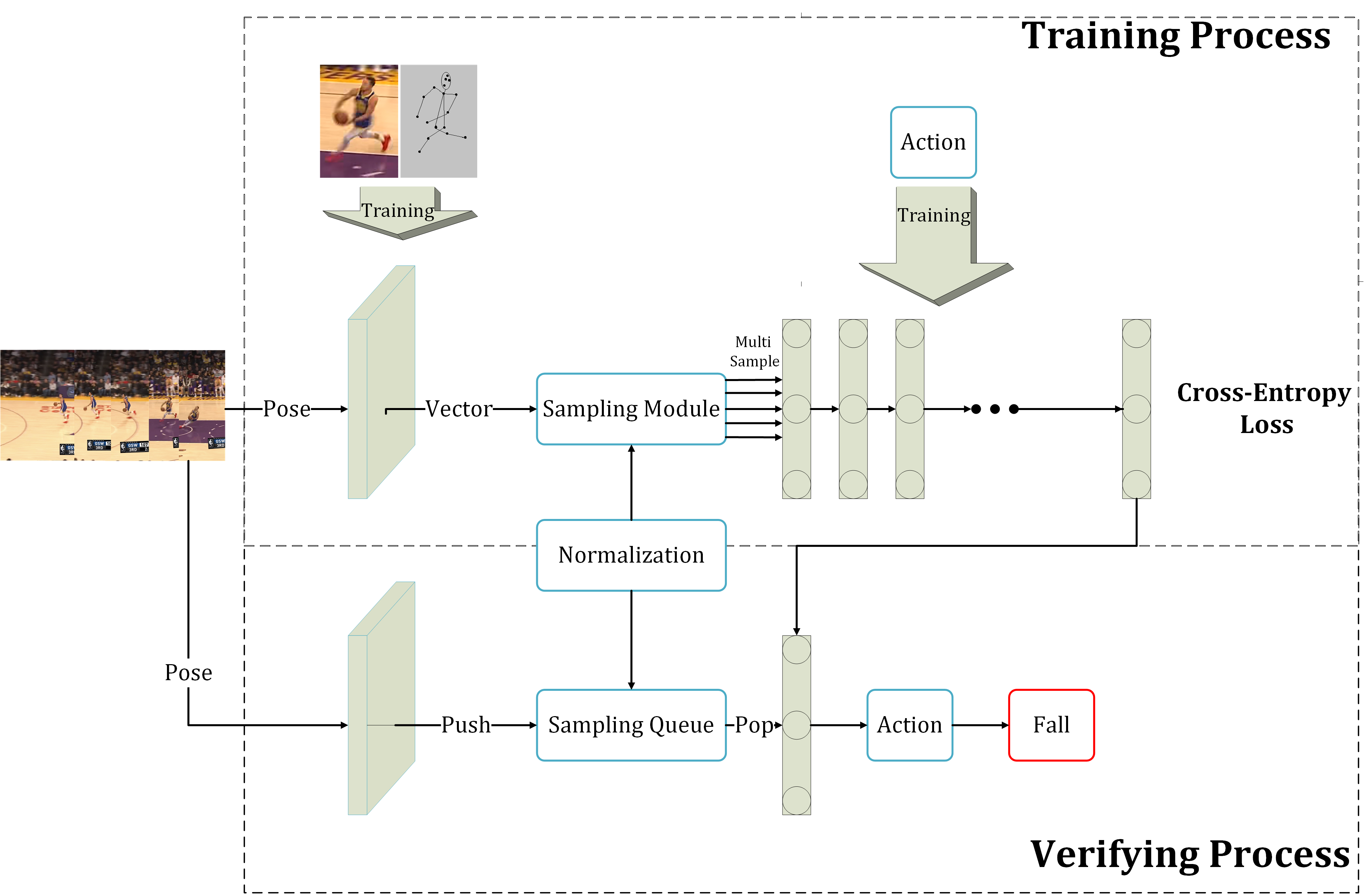}}
\caption{The Whole Model Illustration. The upper area of the figure demonstrates training details of the model and the lower area of the figure demonstrates the verifying detail of the model which would infer the action name. The original input is the video of falling action and the pose model is improved by the single image with its keypoint information. The skeleton image is to illustrate how the keypoint of the image would be marked. In the figure, the interval is set to 5 so 5 sequences are passed to the GRU in the training process.}
\label{fig}
\end{figure*}

This paper will not describe the details of the GRU as demonstrating the detailed function here. We use almost the same structure of GRU as it was presented in the official implementation. The GRU consists of three kinds of gates to learn spatio-temporal information. The reset gate started to update the next iteration and normalize the data. Our activation function $\varphi(\cdot)$ here would take effect. Finally, the unique update gate for all the units would update the stored information. After adequate iterations, we would calculate our expecting result.

\subsection{Our design}
This subsection would introduce the structure of our model. Broadly, taking an input of a sequence of videos, our model would predict the action type. The model consists of two portions: the first one is the pose model and the second one is the GRU model with our unique sampling method and normalization module between these two portions. The pose model would extract the keypoints to convert the two-dimension video into a one-dimension sequence. After the handle of the sampling module, the several times larger augmented sequence would be generated in the sampling module. Finally, the action would be predicted according to the sequences. The details of the model are elaborated in the following paragraph.\\

The training process is shown in figure 3. For the training process, the video would first be transferred into human keypoint information. The groundtruth keypoints of the single human image would supervise the pose model. Obviously, not all the keypoints are available here because some of the keypoints may not exist in the image, and we would note that $n_{k}$ as the number of the keypoints that we have detected, and $(x_{i},y_{i})$ as each pair of the keypoint with $i$ from 0-17. To be noticed, the keypoints that are not available here would be assigned as $(0,0)$. Then the normalization and sampling would take effect here. \\

The normalization method works on each pair of the body keypoints $(x_{i},y_{i})$. The normalization module concentrates on two tasks -- one is globalization and the other is standardization. The globalization process begins with selecting an initial point to replace the original point $(0,0)$, which can eliminate the space interference factor. The new original point is selected by computing the average of the valid keypoints by the formula:\\

\begin{equation}
\begin{aligned}
    \overline{x} = \frac{\sum_{i=0}^{17}{x_{i}}}{n_{k}}
\end{aligned}
\end{equation}
\begin{equation}
    \overline{y} = \frac{\sum_{i=0}^{17}{y_{i}}}{n_{k}}
\end{equation}

The $(\overline{x}, \overline{y})$ would be the new original point. Then the standardization process starts. Assuming the width of the image is w and the height of the image is h. Consequently, each point would be updated by:\\

\begin{equation}
    x_{i} = \frac{(x_{i} - \overline{x})}{w}
\end{equation}
\begin{equation}
    y_{i} = \frac{(y_{i} - \overline{y})}{h}
\end{equation}
The generating new pair $(x_{i},y_{i})$ would range from (-1,1). The standardization puts all the points in the same scope and would make our model more robust.\\

Then the sampling module would convert the video keypoint vectors into multiple keypoint sequences according to the number of intervals, each of which contains a concrete number of frames' keypoints. For example, 2 intervals mean that the sequences would be generated by sampling the keypoint vectors per 2 vectors, and the original videos would generate 2 sequences. 10 keypoint vectors generated from a video with interval as 2 would be divided into two parts: the 1st, 3rd, 5th, 7th, 9th sequence and the 2nd, 4th, 6th, 8th, 10th sequence. Furthermore, these sequences would then be recognized as the same action as they are generated from the same video. Meanwhile, all these sequences can be recognized as separate which means we make our training dataset much larger. The rest GRU would supervise itself based on its own marked action.\\

The verifying process is shown in figure 3. The verifying process contains a queue-based model that is similar to the function of the sampling module. It shares the identical initial stage -- first, we need the keypoint of the video sequential frames and do the normalization process. Then each frame keypoints would be pushed into a queue with a concrete capacity, which is manually set. Once the queue reached its own capacity, the queue would pop out sequences as the interval it sets. For example, a full queue with 50 capacity and 10 intervals would pop out a sequence with 1st, 11th, 21st, 31st and 41st vector as the output. Then according to the action type and these sequences, the GRU can finish its verifying process. \\

For real-time action recognition, the action result would be immediately output to the screen or other connected devices. However, things can be different for a video. For a video, the model would store all the actions of the sequences and output the action name that is most probably the theme of the video. It can be possible because most of the time, we can abandon the static actions and focus on the longest-lasting dynamic behavior, which means rapid change. \\

\section{Experiments}
\subsection{Data Prepossessing and Division}\label{AA}
All the frames in the video would firstly enter the pretrained yolo-version-three model to predict the bounding box of a single person. As all of our training datasets only contain video with a single person, we would not consider the tracking information of each person. Then the bounding box would be put into the pose model to finish the rest training process.\\

\subsection{Model Training}
This section describes the details of model training according to the method above. For pose model, besides the existed data in the previous dataset, we add other unused public datasets to the pose training process. For each pair of data, we supplement the image with 18 keypoints: two eyes, one nose, one forehead, one neck, two shoulders, one chest, two elbows, two hands, two crotches, two thighs, and two feet. All the keypoint information is marked with JSON File with labelme(https://github.com/wkentaro/labelme) tool at first and transferred into a one-dimension vector.\\

For our GRU\cite{pmlr-v63-gao30, chung2014empirical, 8053243} model, the model takes in a one-dimension vector and output an action name. The whole process is recognized as a classification task. For our model, we use three GRU layers for the model.  The activation function is tanh function. The classification loss function is the cross-entropy function. Four dropout layers are implemented to avoid overfitting.\\

\section{Results and Analysis}
\subsection{Performance on our own dataset}
We compare three kinds of models on our own dataset. The first one is the baseline -- the model with the conventional dense sampling method -- the sampling method is dense here without any interval. The second one is the model with flexible-interval sampling but without normalization. It means that we only deal with the data by dividing the width and height without the new original point selection. And the last one is the model with both flexible-interval sampling and normalization module. It is the model that we describe above.\\

For our designed model, the accuracy is elaborated in figure 4. Figure 4 describes how the loss function and accuracy develops in the whole training process. The whole model becomes steady after 50 epochs.\\

\begin{figure*}[htbp]
\centerline{\includegraphics[width = 19cm]{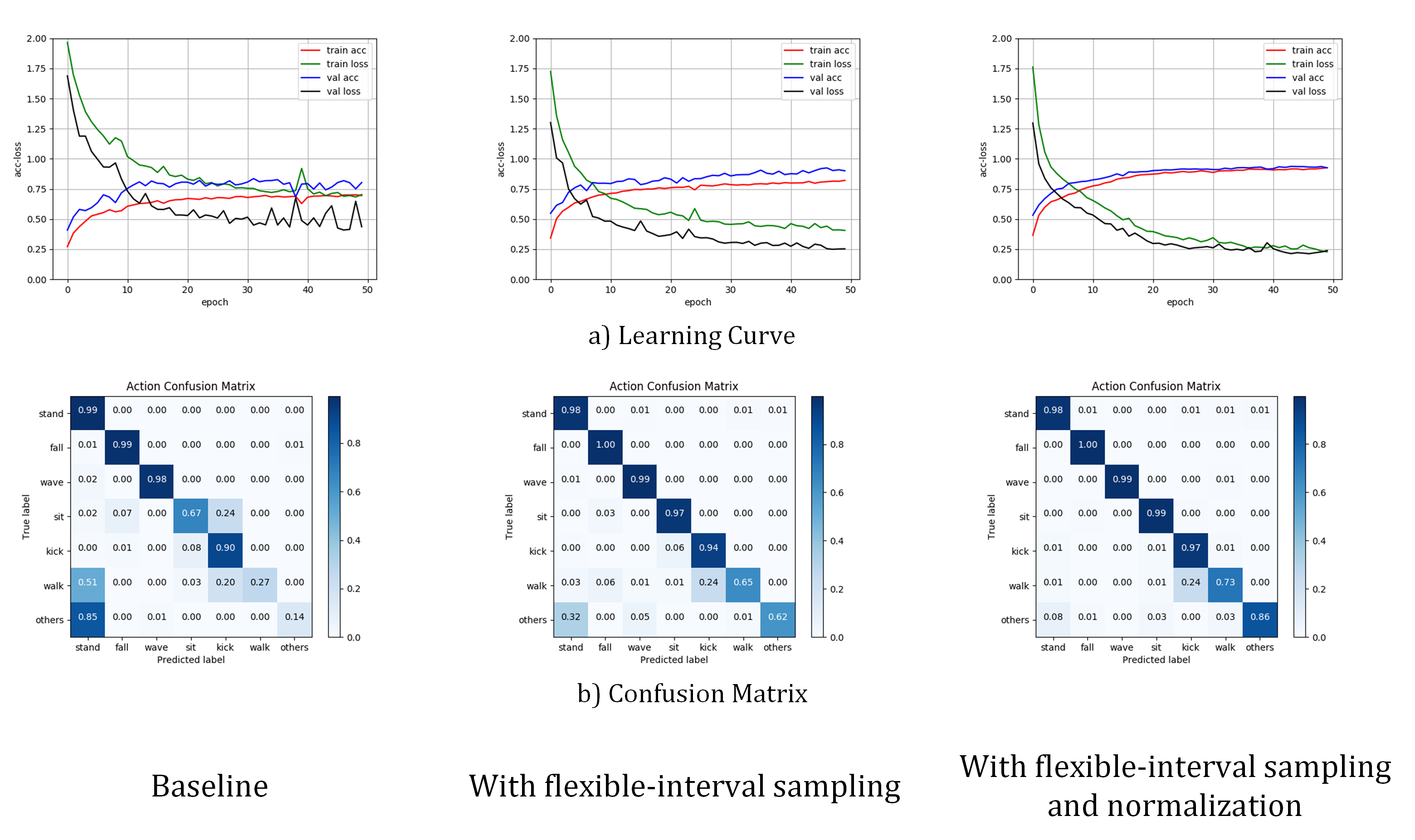}}
\caption{a)Training Result on Our STH-Dataset. The three whole loss vibrates a lot before 50 epochs as the graph shows. The accuracy of the model on our own dataset comes close to 1. Our methods improve the accuracy of the model, especially on the last two categories. b)We compare the confusion matrix of three models on our own dataset. The x-axis denotes the predicted label of the model which is the result of the model. The y-axis denotes the truth label of the model which is the groundtruth. Our method has an accuracy at 92\% with norm and 90\% without norm, and the baseline has an accuracy at 80\%. It is clear to see that the last two methods own better performance. }
\label{fig}
\end{figure*}

Figure 4 compares these three models on the accuracy of the multiple action types, which obviously shows the high precision with flexible-sampling. It can be demonstrated that our flexible-sampling method is pretty effective. However, the promotion by the normalization is not as effective as the previous one. In the last two confusion matrix, the numbers centralize in the diagonal so most of the videos are predicted right. For each action type, the result gets improved. \\

\subsection{Performance on public dataset}
We conduct more experiments to demonstrate our model can still be high-efficient in other similar public action datasets. Specifically, we use the UCLA-Northwestern Action Dataset\cite{Wang_2014_CVPR} and SYSU Dataset\cite{hu2017jointly}. SYSU Dataset provides multiple sequential photos, and we recover videos from these photos. Both of these two datasets hold similar actions, so we select them as our subjects. The details of these datasets are elaborated in the following paragraph. \\

\paragraph{UCLA-Northwestern Action Dataset}
This dataset contains RGB, depth and human skeleton data captured simultaneously by three Kinect cameras. For our task, we only utilize RGB videos. These datasets include 8 action categories. Each action is performed by 10 actors. All of these actions are taken place in the same location field. This dataset contains data taken from a variety of viewpoints.

\paragraph{SYSU Dataset}
The action dataset contains two categories of input: RGB photos and depth photos. The dataset focuses on human-object interaction, which will be made available to the human action benchmarking and analysis. The dataset contains 12 actions and each action is conducted by 40 persons.

For each dataset, we use our pose model\cite{2018_CVPR, cao2018openpose, simon2017hand, cao2017realtime, wei2016cpm,fang2017rmpe,xiu2018poseflow} to infer the keypoints of each video to convert them into one-dimension vectors. We first retrain our model based on these different datasets and then test the accuracy of the model. And the accuracy result on each dataset is demonstrated in figure 5. \\
\begin{figure*}[htbp]
\centerline{\includegraphics[width = 19cm]{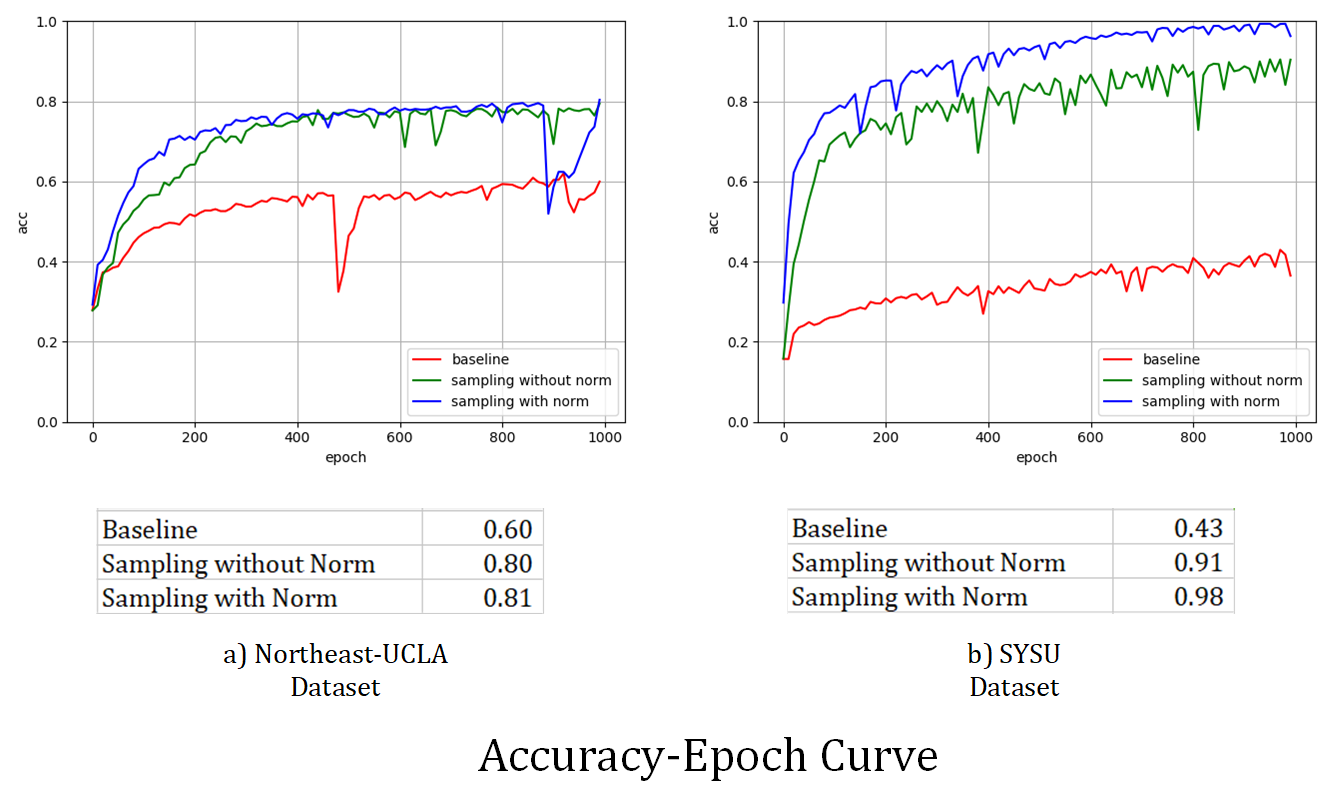}}
\caption{Figure 5 demonstrates the comparison of accuracy between the baseline and our model. Although it seems a little fluctuant at first, the final accuracy becomes stable. The accuracy in the table demonstrates that on both two datasets our improvement is pretty effective.}
\label{fig}
\end{figure*}\\

As figure 5 demonstrates, our model can significantly promote the accuracy of the action recognition task on these two datasets. The promotion of the effect on the UCLA-Northwestern Action Dataset is not obvious because each video contains too few frames. Consequently, we can only choose 2 or 3 as our interval for most of the videos with less than 10 frames. However, the effect is still obvious. For SYSU dataset, it demonstrates the power of the normalization. The human taking action in the SYSU dataset is not always fixed in one place, he or she may take actions in different regions of the videos. Normalization unifies them so that the spatio-temporal network can concentrate on the action itself despite which region the human stays. Some of the other methods utilize the depth figures or skeleton images in the dataset. On the RGB video task, our method owns much better performance than the others. The importance of the RGB video has already been placed in the introduction.

\section{Conclusions and Future Work}
In this work, we proposed a novel spatio-temporal human action with flexible-interval sampling and normalization by taking advantage of pose model and GRU. To improve the accuracy of the current model and make the current model more robust, we propose a novel model with sampling and normalization. To take sufficient advantage of all the training dataset, our own-made sampling module helps to create more training data and help the model understand more what the real behavior is. Finally, we combine the existed pose model, sampling module and GRU to finally generate with a cross-entropy loss function. Experimental result demonstrates that our model can classify the action accurately despite the similar background.\\

\bibliographystyle{plain}
\bibliography{ref}

\end{document}